\documentclass[journal]{IEEEtran}
\usepackage{times}
\usepackage{epsfig}
\usepackage{graphicx}
\usepackage{amsmath}
\usepackage{amssymb}
\usepackage{booktabs}
\usepackage{multirow}
\usepackage{nicematrix}
\usepackage{etoolbox}
\makeatletter
\patchcmd{\@makecaption}
  {\scshape}
  {}
  {}
  {}
\makeatother

\usepackage{booktabs}
\usepackage{array, caption, threeparttable}
\ifCLASSINFOpdf
\else
\fi
\begin{document}
%
\title{Geometry-Entangled Visual Semantic Transformer for Image Captioning}
%
%
%

\author{Ling Cheng,
        Wei Wei,
        Feida Zhu,
        Yong Liu,~\IEEEmembership{Member,~IEEE,}
        and~Chunyan Miao,~\IEEEmembership{Senior Member,~IEEE}
\thanks{This work was supported in part by the National Natural Science Foundation of China under Grant 61602197}
}

%
%

\markboth{Journal of \LaTeX\ Class Files,~Vol.~14, No.~8, August~2015}%
{Shell \MakeLowercase{\textit{et al.}}: Bare Demo of IEEEtran.cls for IEEE Journals}
%



\maketitle

\begin{abstract}
	Recent advancements of image captioning have featured Visual-Semantic Fusion or Geometry-Aid attention refinement.
	However, those fusion-based models, they are still criticized for the lack of geometry information for inter and intra attention refinement.
	On the other side, models based on Geometry-Aid attention still suffer from the modality gap between visual and semantic information.
	In this paper, we introduce a novel Geometry-Entangled Visual Semantic Transformer (GEVST) network 
	to realize the complementary advantages of Visual-Semantic Fusion and Geometry-Aid attention refinement.
	Concretely, a Dense-Cap model proposes some dense captions with corresponding geometry information at first.
	Then, to empower GEVST with the ability to bridge the modality gap among visual and semantic information,
	we build four parallel transformer encoders VV(Pure Visual), VS(Semantic fused to Visual), SV(Visual fused to Semantic), SS(Pure Semantic) for final caption generation.
	Both visual and semantic geometry features 
	are used in the Fusion module and also the Self-Attention module for better attention measurement.
	To validate our model, we conduct extensive experiments on the MS-COCO dataset,
	the experimental results show that our GEVST model can obtain promising performance gains.
\end{abstract}

\IEEEpeerreviewmaketitle

\section{Introduction}
\label{sec:Introduction}

%
Recent advances in deep learning have resulted in great progress in both the computer vision and natural language processing communities. These achievements make it possible to connect vision and language, and facilitate multi-modal learning tasks such as image-text matching, visual question answering, visual grounding and image captioning. Image captioning aims to automatically describe an image’s content using a natural language sentence. The task is challenging since it requires one to recognize key objects in an image, and to understand their relationships with each other. Inspired by machine translation, encoder-decoder architecture became the most successful approach for image caption task. Encoder encodes the visual(object region or grid visual feature) or semantic(attribute words) information into feature vectors. Given the input features, decoder iteratively generates the output caption words. Based on this architecture, various improvements have been applied into encoder, decoder and training method to further improve the performance. 

Previously, there are three kinds of method to improve the model.
The first one is based on attention mechanism. To establish the fine-grained connections of caption words and their related image regions, attention can be seamlessly inserted into the framework. Then, a lot of works focus on improving attention measurement to enhance the interaction between visual content and natural sentence to boost image captioning. The second one treat image captioning as a multi-modal problem, by gradually fusing visual and semantic information or exploiting semantic and visual information simultaneously, they can identify the equivalent visual signals especially when predicting highly abstract words. The last one is aim to address the exposure bias of generated captions by using the cross-entropy loss, reinforcement learning (RL)-based algorithm~\cite{E_1, E_2,E_3} are designed to directly optimize the non-differentiable evaluation metrics (e.g., BLEU, METEOR, ROUGE, CIDEr and SPICE~\cite{F_1, F_2, F_3, F_4, F_5}).

Despite those reinforcement learning based algorithms which are compatible to different models, there existing two main limitations in both attention focused mechanism and multi-modal focused mechanism:
1) Even with visual sentinel mechanism, typical attention focused mechanisms are still incapable of inter-modal interactions, thus they are arduous to identify the equivalent visual signals especially when predicting highly abstract words.
2) Multi-modal focused models typically fuse visual branch(object region or grid visual feature) and semantic branch(attribute words) with concatenation or summation, these method are usually shallow and may fail to fully understand the complex relationships among information from two different modalities. Besides, even attention mechanism are used in each branch, since these primitive semantic features have no geometry information as guidance, noise from different modal will deteriorate the attention measurement.

To address the limitation in attention focused models, we extend the Transformer model for machine translation to a Visual-Semantic Transformer model for image captioning. Different from previous Transformer based captioning models, there are four parallel branches in the encoder (VV:Pure Visual, VS:Semantic fused to Visual, SV:Visual fused to Semantic, SS:Pure Semantic). Different branch stands for different ratios of visual information to semantic information. By dynamically adjust each branch's contribution at different time step, model can not only maintain sophisticated attention measurement in each branch, but can also generate words with abstract semantic meaning.
To address the limitation in multi-modal focused models, rather than using MIL model to extract attribute words, we use Dense-Caption model to extract dense captions which describe sub-images and each dense caption has a bounding box. By properly taking both content and geometry information into a fusion module, model can not only encode this complex multi-modal relationship to a deeper degree, but can also generate inter geometry features between two different modals. Then, inter and intra geometry information will contribute to a better attention measurement in each self-attention encoder branch. Thus we call the model as Geometry-Entangled Visual-Semantic Transformer(GEVST).

To summarize, the main contributions of this study are three-fold:
\begin{itemize}
	\item Inter- and intra-geometry and content relation in both fusion and self-attention is first proposed in the GEVST model. Our model is not only capable of fusing two modality information based on both content and geometry information, by feeding both inter and intra geometry features into self-attention module, model can also get a better attention measurement which improves image captioning performance.
	\item To dynamically give a more fine-grained ratio and higher order interaction between visual and semantic, we perform a cross-attention with 4 encoding branches, these branches' outputs with multi-VS ratio are then summed together after modulation. 
	\item Extensive experiments on the benchmark \textsf{MSCOCO}~\cite{G_1, G_2} image captioning dataset are conducted to quantitatively and qualitatively prove the effectiveness of the proposed models. The experimental results show that the GEVST significantly outperforms previous state-of-the-art approaches with a single model.	
\end{itemize}

The rest of the paper is organized as follows: 
In section II, we review the related work of image captioning approaches, especially those attention focused and multi-modal focused mechanisms. 
In section III, we introduce our visual-semantic fusion module which take both content and geometry information into consideration.
In section IV, we introduce our Geomerty-Entangled self-attention module, and will also elaborate the architecture of four parallel branches in our GEVST model.
In section V, we introduce our extensive experimental results for evaluation and use the benchmark MSCOCO image captioning dataset to evaluate our proposed approaches. 
Finally, we conclude this work in section VI.

\section{Related Work}
\label{sec:Relatedwork}
In this section, we briefly review the most relevant research on image captioning, especially those attention focused and multi-modal focused mechanisms.

\subsection{Image Captioning}
Previously, the research on image captioning can be mainly divided into three categories: template-based approaches~\cite{A_1, A_2}, retrieval-based approaches~\cite{A_3, A_4, A_5}, and generation-based approaches~\cite{A_6, A_7, A_8, A_10, A_11}. 
The template-based approaches address the task using a two-stage strategy, it first detects the objects, their attributes, and object relationships in an image, and then fills the extracted information in a pre-designed sentence template. Both Conditional Random Field (CRF) and Hidden Markov Model (HMM)~\cite{A_12, A_13} are used to predict labels based on the detected objects, attributes, and prepositions, and then generate caption sentences with a template. Obviously, those captions generated by the template based approaches are highly depend on the quality of the templates and usually follow the syntactical structures.  However, the diversity of the generated captions is severely restricted, as a result, the generated captions are often rigid, and lack naturalness.
To ease the diversity problem, is to directly retrieve the target sentences for the query image from a large-scale caption database with respect to their cross-modal similarities in a multi-modal embedding space. The core idea~\cite{A_3} is to propose an approach to match the image-caption pairs based on the alignment of visual segments and caption segments. For prediction, the cross-modal matching over the whole caption database is performed to generate the caption for one image. Further improvements~\cite{A_4, A_5} are focusing on different metrics or loss functions to learn the cross-modal matching model. However, the result captions are fixed, and may fail to describe the object combinations in a new image. To increase caption's diversity, caption database has to be as large as possible. Whereas, for large database, there is a big problem in retrieval efficiency. Thus, the diversity problem has not been completely resolved.
Different from template-based and retrieval-based models, in recent years, With the successes of \emph{sequence generation} in machine translation~\cite{A_14, A_15}, generation-based model has emerged as a promising solution to image captioning. In this kind of architecture, model tries to build an end-to-end system with encoder-decoder framework, thus, it can generate novel captions with more flexible syntactical structures. Vinyals et al. propose an encoder(CNN Based)-decoder(LSTM Based) architecture as its backbones. After that, this kind of encoder-decoder architecture became the mainstream for generation-based approaches. Similar architectures are also proposed by Donahue et al.~\cite{A_16} and Karpathy et al.~\cite{A_11}. Due to the flexibility and excellent performance, most of subsequent improvements are based on the generation-based models. 

\subsection{Attention Focused Model}
Within the encoder-decoder framework, one of the most important improvements for generation-based models is the attention mechanism. Xu et al.~\cite{A_6} proposed soft/hard attention mechanism to automatically focus
on the salient regions when generating corresponding words. The attention model is a pluggable module that can be seamlessly inserted into previous approaches to remarkably improve the caption quality. 
The attention model is further improved in~\cite{B_1, B_2, A_8, B_3, B_4, B_6, B_7, B_8, B_10}. 
Lu et al. presented an adaptive attention encoder-decoder model for automatically deciding when to rely on visual or language signals.
Chen et al. proposed a spatial- and channel-wise attention model to attend to visual features.
Anderson et al. introduced a bottom-up module, that uses a pre-trained object detector to extract region-based image features, and a top-down module that utilizes soft attention to dynamically attend to these object 
Gu et al. adopted multi-stage learning for coarse-to-fine image captioning by stacking three layers of LSTMs, where each layer of LSTM was different from other layers of LSTMs.
Zhou et al. designed a novel salience-enhanced re-captioning framework via two-phase learning is developed for the optimization of image caption generation
Huang et al. proposed attention on attention module enhances visual attention by further measuring the relevance between the attention result and the query.
Wang et al. proposed a hierarchical attention network to combine text, grids, and regions with a relation module to exploit the inherent relationship among diverse features.
Herdade et al. utilized bounding boxes of regions to model location relationships between regions in a relative manner.
Cornia et al. presented a mesh-like structure transformer to exploit both low-level and high-level contributions.
Through all of these works, it is confirmed that improving attention measurement is an effective way to enhance the interaction between visual content and natural sentence and thus boost image captioning.
However, typical attention mechanisms are arduous to identify the equivalent visual signals especially when predicting highly abstract words. This phenomenon is known as the semantic gap between vision and language. 

\subsection{Multi-Modal Focused Model}
This problem in those attention focused model can be overcome by providing semantic attributes that are homologous to language~\cite{C_1, A_7, C_2, C_3, C_5, C_6, C_7}.
Hao et al. use multiple instance learning to train to detect words(nouns, verbs, and adjectives) that commonly occur in captions.
Based on the these detected words, You et al. leveraged the high-level semantic information directly with semantic attention.
As it was justified that the combination of the two complementary attention paradigms can alleviate the harmful impacts of the semantic gap. Therefor, Li et al. proposed a two-layered LSTM that the visual and semantic
attentions are separately conducted at each layer.
Yu et al. designed a multi-modal transformer model which simultaneously captures intra- and inter-modal interactions in a unified attention block.
Huang et al. also introduced two modules to couple attribute detection with image captioning as well as prompt attributes by predicting appropriate attributes at each time step.
Another way to bridge the modality gap is to employ graph convolutional neural networks(GCN).
Yao et al. built a GCN model to explore the spatial and semantic relationships. Then, by late fusion, they combined two LSTM language models that are independently trained on different modalities.
Similarly, Xu et al. used a multi-modal graph convolution network to modulate scene graphs into visual representations.
Whereas, the interactions between different modals are usually shallow and may fail to fully understand the complex relationships among these two modal information. Besides, even attention mechanism are used in each branch, since their semantic features have no geometry information as guidance.

\subsection{Caption Related Task}
Dense-caption task is also a task aimed at caption generation. Dense-caption performs intensive captions\cite{D_1, D_2, D_3} on the image. Beside giving caption which describes a certain sub-image, it can also give the geometry information of this caption. And since the output is a complete sentence rather than attribute words, it can provide more advanced semantic and linguistic information. Like image-caption, Visual Question Answering(VQA) also aim at handling the modal-crossing problem. In VQA, although attention mechanism helps model to focus on the visual content relevant to the question, it's still insufficient to model complex reasoning features. Rather than simple attention mechanism, multi-modal fusion plays the key roles in better performance. In VQA, there are many visual semantic fusion methods\cite{D_4, D_5, D_6, D_7} are proved to be more powerful than just concatenation or element-wise addition.

\section{Proposed Method}
\label{sec:Methodology}
\begin{figure*}[!t]
    \includegraphics[width=2\columnwidth, angle=0]{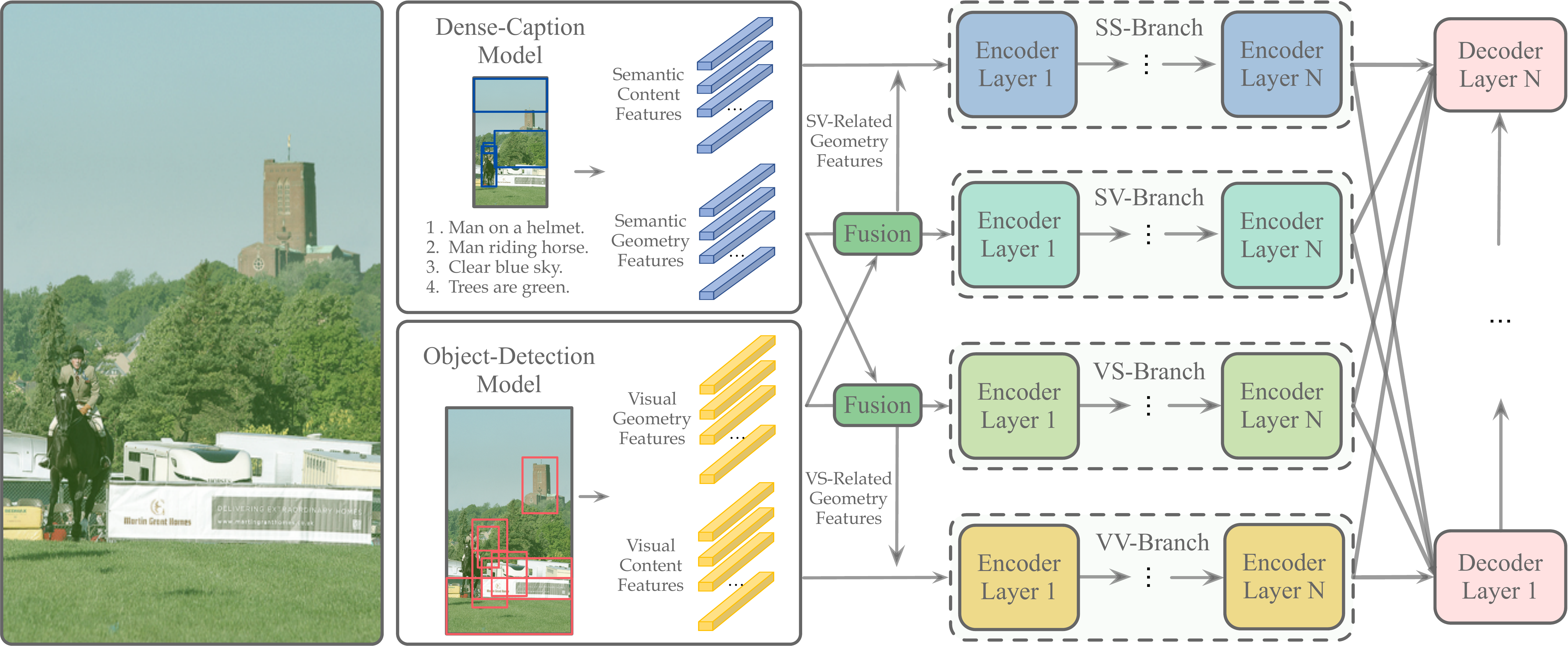}
    \vspace{-1ex}
	\caption{Overview of our proposed GEVST approach.Firstly, Visual encoder and semantic encoder will generate objects region feature and dense-captions with corresponding geometry information.
	After a fusion model, we can get the related geometry features between two modalities. Decoder is to generate the output caption based on the output from 4 encoder branches.}
	\label{Model_Pipeline}
	\vspace{-0.0cm}
\end{figure*}

In this section, we first briefly describe the preliminary knowledge of the Transformer model\cite{D_8}, 
After that, we will elaborate the data preparation.
Then, we introduce the Geometry-Entangled Visual Semantic Transformer(GEVST) framework as shown in the Fig \ref{Model_Pipeline}, 
which can be conceptually divided into a fusion model, an encoder and a decoder module, 
both encoder and decoder made of stacks of attentive layers.
The fusion model fuses visual and semantic information based on the content and geometry information.
Encoder is in charge of processing deep representation of the input information in a self-attention manner. 
The decoder is to generate the output caption word by word.
\vspace{1mm}
\subsection{The Transformer Model}
\label{sec:The_Transformer_Model}
The Transformer model was first proposed for machine translation, and has been successfully applied to many natural language processing tasks. 
All intra- and inter-modality interactions between word and image-level features are modeled via scaled dot-product attention, without using recurrence.
Attention operates on three sets of vectors, namely a set of queries ${Q}$, keys ${K}$ and values ${V}$, 
and takes a weighted sum of value vectors according to a similarity distribution between query and key vectors. 
In the case of scaled dot-product attention, the operator can be formally defined as below,
\begin{equation}
\label{eq:scaled dot-product attention}
{F} = Attention({Q}, {K}, {V}) = softmax(\frac{{Q}{K}^T}{\sqrt{d}}){V},
\end{equation}
where ${F}$ is a matrix of $n_q$ attended query vectors, ${Q}$ is a matrix of $n_q$ query vectors, ${K}$ and ${V}$ both contain $n_k$ keys and values, all with the same dimension, and $d$ is a scaling factor. 
Instead of performing a single attention function for the queries, 
multi-head attention is to allow the model to attend to diverse information from different representation sub-spaces. 
The multi-head attention contains $h$ parallel ‘heads’ with each head corresponding to an independent scaled dot-product attention function.  The attended features ${F}$ of the multi-head attention functions is given as follows:
\begin{equation}
\begin{aligned}
\label{eq:multi-head-attention}
& {F} = MultiHead({Q}, {K}, {V}) = Concat({H_1},\cdots,{H_h})W^O \\
& {H_i} = Attention({Q}{W^Q_i}, {K}{W^K_i}, {V}{W^V_i})
\end{aligned}
\end{equation}
where ${W^Q_i}, {W^K_i}, {W^V_i} \in \mathbb{R}^{{d}\times\frac{d}{h}}$, are the independent head projection matrices.
${W^O_i} \in \mathbb{R}^{{d}\times{d}}$ denotes the linear transformation. 
Note that the bias terms in linear layers are omitted for the sake of concise expression,
and the subsequent descriptions follow the same principle.

The Transformer is a deep end-to-end architecture that stacks attention blocks to form an encoder-decoder strategy.
Both the encoder and the decoder consist of $N$ attention block, 
and each attention block contains the multi-head attention modules. 
The multi-head attention module learns the attended features that consider the pairwise interactions between two input features,
In the encoder, each attention block is self-attentional such that the queries, keys and values refer to the same input features. 
In the decoder contains a self-attention layer and a guided attention layer. 
It first models the self-attention of given input features and then takes the output features of the last encoder attention block to guide the attention learning. 

\subsection{Data Preparation}
\label{sec:Data_Preparation}
As mentioned in Section~\ref{sec:Introduction},
our model is based on both the visual representation and semantic representation.
In order to model the given image in visual-level, each element $\vec{v_i}$ of visual feature vector ${V}$ is defined as the mean-pooled convolutional feature of the $i$-th salient region, which is a $d_v$-dimensional fixed-size feature vector (empirically set as $2048$). These visual features are extracted by means of the output of final convolution layer of \emph{Faster-RCNN} model~\cite{A_8}, which is widely used in previous work~\cite{B_4, B_6, B_7, B_8}.
Thus, for the visual information, we have ${V}=\{\vec{v}_{i}\}^{N_v}_{i=1}~(\vec{v}_{i}\in \mathbb{R}^{d_v})$ to denote the $N_v$ \emph{visual} feature vectors with $d_v$-dimension.
To fetch semantic representation for the given image, we consider to employ dense-caption model to obtain a list of dense-captions that are most likely to appear in the image.
Same to the work~\cite{D_1}, for each image, we get a list of dense-captions, corresponding confidence scores and bounding boxes information.
Then, each dense-caption will be encoded by a standard $3$-layers transformer encoder. Then, the confidence scores are used to assign a weight to corresponding phrases. 
At last, we get a set of $d_s$-dimensional fixed-size semantic feature vectors, the $i$-th vector is denoted as ($\vec{s_i}$).
Thus now, we have ${S}=\{\vec{s}_i\}^{N_s}_{i=1}~(\vec{s}_{i}\in \mathbb{R}^{d_s})$ to denote the $N_s$ \emph{semantic} dense-caption vectors with $d_s$-dimension.
The inherent geometric structure among the input visual objects and semantic objects is beneficial for reasoning,
thus bounding box coordinates for both visual and semantic features are feed into the model.

\subsection{Visual-Semantic Fusion}
\label{sec:Visual_Semantic_Fusion}
\begin{figure*}[!t]
	\includegraphics[width=2\columnwidth, angle=0]{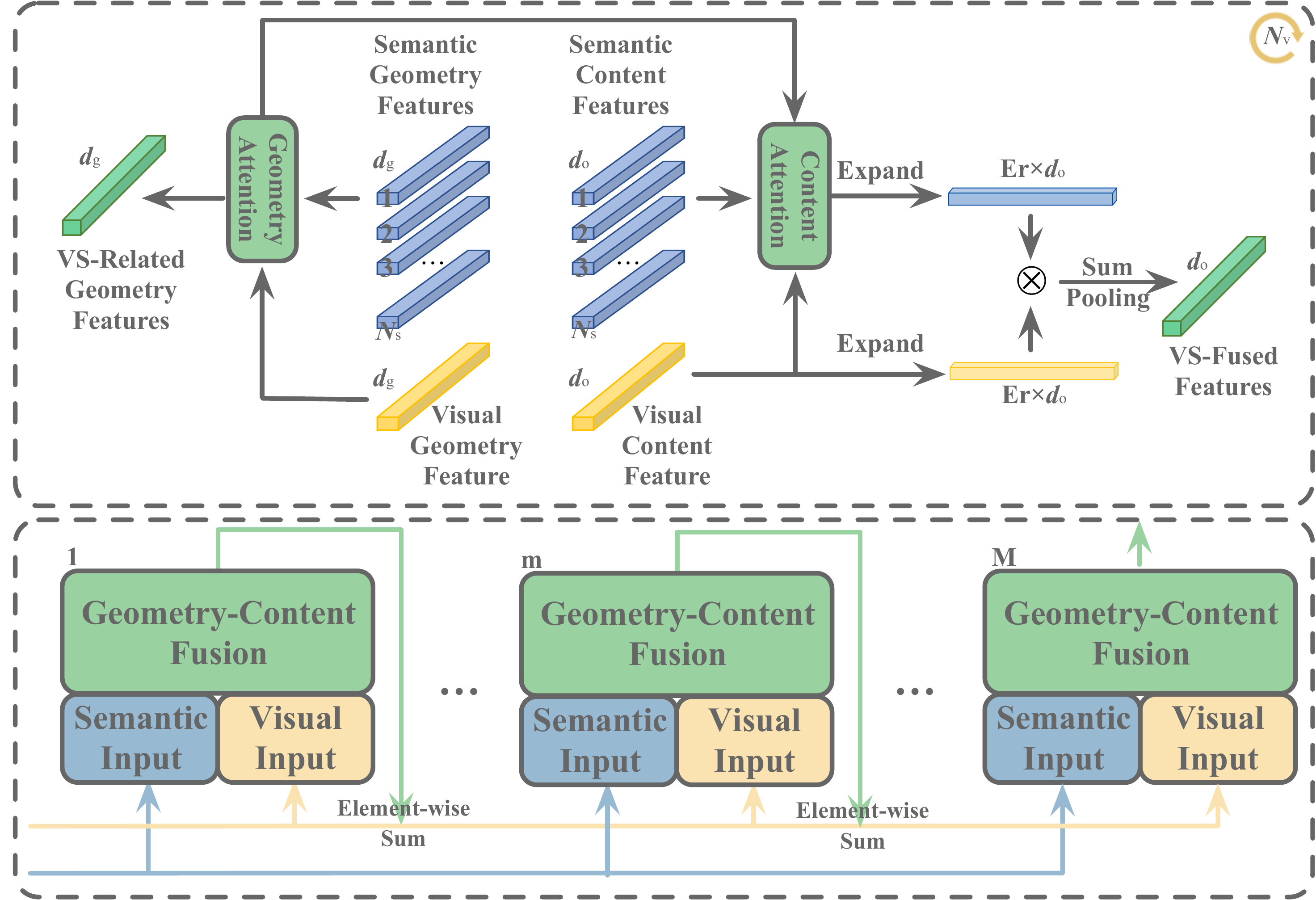}
	\vspace{-1ex}
	\caption{Geometry-Content Fusion Cell(above) and Stack fusion module(below) for $VS$-Branch, where the semantic information is fused into visual information. 
	For each visual feature$\{\vec{v}_{i}\}^{N_v}_{i=1}$, the attention weight of Geometry-Attention and Content-Attention will be summed to generate the final attend-semantic feature.
	Besides, the weighted-summation of semantic geometry feature is used to represent the intra-geometry information for the given visual feature.}
	\label{fig:GEF_Model}
	\vspace{-0.0cm}
\end{figure*}

\noindent\textbf{Geometry-Content Fusion Cell}
For a given visual feature, it has different relationship with different dense-captions.
The relation can be divided into two types, the content relation and the geometry relation.
Thus, to fuse features from different modalities, we propose a Stack Geometry-Content Fusion.
For example, when fusing the semantic information into the visual information(VS-Branch),
both visual and semantic bounding box vectors will be projected into high dimension representations ${V}^{geo}$ and ${S}^{geo}$
as shown in the Fig \ref{fig:GEF_Model}, 
each fusion cell contains a geometry attention layer and a content attention layer to encode two kinds of relation simultaneously.
To select relevant semantic features for the $i$-th visual content feature ${V}^{con}_i$, 
both content and geometry attention layer will output an attention map,
then, two maps are added together for generating the weighted sum semantic content feature $\hat{S}_i$.
%
\begin{equation}
{a}^{con}_{i,j}= {W^{con}}\mbox{tanh}({{W^{con}_{s}}{S}^{con}_j} + {{W^{con}_{v}}{V}^{con}_i})
\end{equation}
\begin{equation}
{a}^{geo}_{i,j}= {W^{geo}}\mbox{tanh}({{W^{geo}_{s}}{S}^{geo}_j} + {{W^{geo}_{v}}{V}^{geo}_i})
\end{equation}
\begin{equation}
({\alpha}^{con}_{i,1}, \cdots,  {\alpha}^{con}_{i,N_s})= \mbox{Softmax}({a}^{con}_{i,1}, \cdots, {a}^{con}_{i,N_s})
\end{equation}
\begin{equation}
({\alpha}^{geo}_{i,1}, \cdots,  {\alpha}^{geo}_{i,N_s})= \mbox{Softmax}({a}^{geo}_{i,1}, \cdots, {a}^{geo}_{i,N_s})
\end{equation}
\begin{equation}
\hat{S}^{con}_i = \sum_{j=1}^{N_s}({\alpha}^{con}_{i,j}+{\alpha}^{geo}_{i,j}){S}^{con}_j
\end{equation}

Then, both the $i$-th visual content feature and the attended semantic content feature will be expanded to high-dimensional representations $\dot{V}_i$ and $\dot{S}_i$.
After element-wise multiplication and a sum pooling function with stride equals to expansion rate $Er$, 
we get the $i$-th Content-based fused feature.
%
\begin{align}
&{\dot{V}_i}= {W_{v}^{Exp}}{V}^{con}_i \\
&{\dot{S}_i}= {W_{s}^{Exp}}\hat{S}^{con}_i
\end{align}
%
\begin{equation}
{{F}_{i}}= \mbox{Sum Pooling}(\dot{S}_i \otimes {\dot{V}_i})
\end{equation}
where, ${W_{v}^{Exp}} \in \mathbb{R}^{Er*d_o \times d_o}$ and ${W_{s}^{Exp}} \in \mathbb{R}^{Er*d_o \times d_o}$ 
are the expansion matrices for visual and semantic branches.

Notice that, the geometry attention weight map can provide the geometry relation between two modalities,
thus, the weighted sum of semantic geometry features $\hat{S}^{geo}_i$ can be seen as inter-geometry features ${V}^{inter}_i$ for the given visual feature to describe the geometry relation between two modalities. 
\begin{equation}
{V}^{inter}_i = \hat{S}^{geo}_i = \sum_{j=1}^{N_s}{\alpha}^{geo}_{i,j}{S}^{geo}_j
\end{equation}
For clarity and consistency in the next sections, the visual geometry features ${V}^{geo}_i$ used previously are denoted as ${V}^{intra}_i$, which stands for the geometry features of the original modality.
Thus, for a visual object, it has a content feature ${V}^{con}_i$, an intra-geometry feature ${V}^{intra}_i$ and an inter-geometry feature ${V}^{inter}_i$.
And it is similarly to every semantic object.

\noindent\textbf{Stack Geometry-Content Fusion}.
To strengthen the fusion effect, a stack structure is proposed as shown in the Fig \ref{fig:GEF_Model}.
Still, take $VS$-Branch for example, to prevent gradients from vanishing or explosion, the fusion result will be skip-connected to the input visual features,
the updated visual features and the original semantic features will be feed to the next cell for further fusion.
And similarly, for $SV$-Branch, the fusion result will be skip-connected to former semantic features, 
the updated semantic features and the original visual features will be feed to the next fusion cell.

\subsection{Geometry-Entangled Self-Attention}
\label{sec:Geometry_Entangled_Attention}
Except from content-based relationships among the input features,
for image captioning, geometric structures are also beneficial for reasoning the interaction.
Former works can only use intra-geometry relation, whereas, 
the important inter-geometry relation between two modalities are missed.
Therefore, we propose a Geometry-Entangled Self-Attention operator. 
In our proposal, along with the content information, two kinds of extra geometry information are feed into Self-Attention layer,
namely, intra- and inter-geometry relation.
Thus, we have three attention maps for the final self-attention measurement.
\begin{figure}
    \includegraphics[width=1\columnwidth, angle=0]{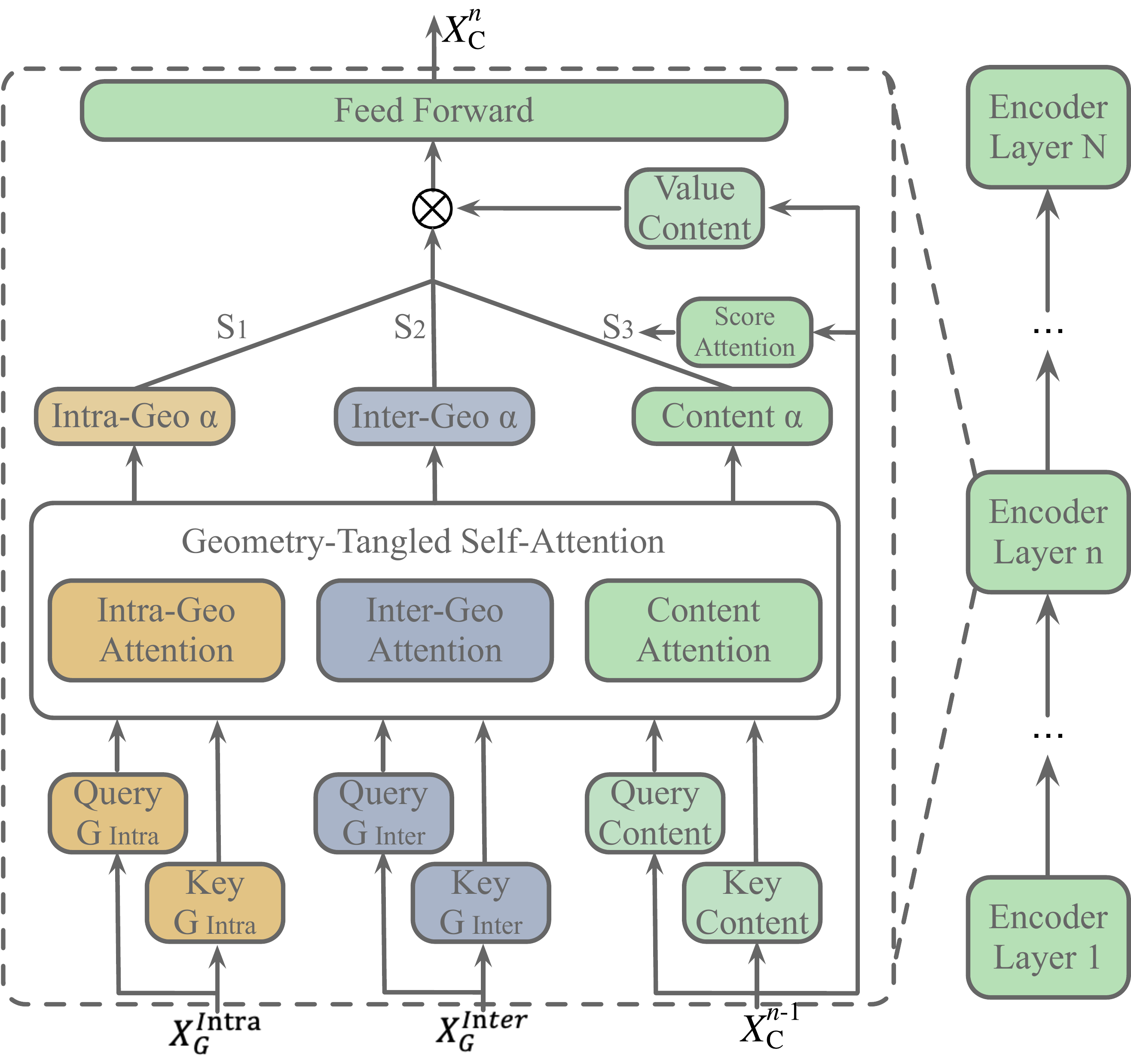}
	\vspace{-1ex}
	\caption{A schematic diagram of Geometry-Entangled Self-Attention layer based on content and geometry interactions.}
	\label{fig:Geometry-Entangled Self-Attention(GESA)}
\end{figure}

As shown in the Fig\ref{fig:Geometry-Entangled Self-Attention(GESA)}, in each encoder layer, 
three sets of keys and queries are used for content, intra- and inter-geometry self-attention maps.
For the content self-attention map which is same to the standard self-attention,
\begin{equation}
\mathit{ATT_1} = \mathit{Softmax}(\frac{{Q_C}{K_C}^T}{\sqrt{d}})
\end{equation}
\begin{equation}
{Q_C} = {X_C^{n-1}}{W^Q_C}
\end{equation}
\begin{equation}
{K_C} = {X_C^{n-1}}{W^K_C}
\end{equation}
where ${X_C^{n-1}}$ is the output values of $n-1$ th encoder layer, ${X_C^{0}}$ is equal to input content features at the first layer. ${W^Q_C}, {W^K_C}$ are the projection matrices for queries and keys, $d$ is the common dimension of all the inputs features,
$ATT_1 \in \mathbb{R}^{N_v \times N_v}$ is the content attention map.
For the intra- and inter-geometry self-attention maps, ${V}^{intra}$ and ${V}^{inter}$ are used to generate intra- and inter-geometry queries and keys.
\begin{equation}
\mathit{ATT_2} = \mathit{Softmax}(\frac{{Q_{G-intra}}{K_{G-intra}}^T}{\sqrt{d}})
\end{equation}
\begin{equation}
\mathit{ATT_3} = \mathit{Softmax}(\frac{{Q_{G-inter}}{K_{G-inter}}^T}{\sqrt{d}})
\end{equation}
\begin{equation}
{Q_{G-intra}} = {{X}^{intra}_G}{W^Q_{G-intra}}
\end{equation}
\begin{equation}
{Q_{G-inter}} = {{X}^{inter}_G}{W^Q_{G-inter}}
\end{equation}
\begin{equation}
{K_{G-intra}} = {{X}^{intra}_G}{W^K_{G-intra}}
\end{equation}
\begin{equation}
{K_{G-inter}} = {{X}^{inter}_G}{W^K_{G-inter}}
\end{equation}
Through intra-geometry queries matrix ${W^Q_{G-intra(inter)}}$ and 
inter-geometry queries matrix ${W^K_{G-intra(inter)}}$, 
intra-geometry input ${X}^{intra}_G$ and inter-geometry input ${X}^{inter}_G$ will generate 
the intra-geometry attention map $ATT_2 \in \mathbb{R}^{N_v \times N_v}$ and 
the inter-geometry attention map $ATT_3 \in \mathbb{R}^{N_v \times N_v}$.

Lastly, to dynamically focus on different self-attention maps at different layers, 
The mean of ${X_C^{n-1}}$ will be feed into a softmax layer to generate 3 scores $c_1, c_2, c_3$ for each self-attention map.
Then, the final attended values are calculated as below,

\begin{equation}
{X_C^{n}} = (\sum_{i=1}^{3}\mathit{ATT_i}\times{c_i}){X_C^{n-1}{W^V_C}}
\end{equation}
\begin{equation}
[c_1, c_2, c_3] = \mathit{Softmax}(\mathit{mean}({X_C^{n-1}}{W^o}))
\end{equation}
where, ${W^V_C}, {W^o}$ are the projection matrices for values update and attention map score generation.

\subsection{Multi-Branches Encoder-Decoder}
\label{sec:Multi_Branches_Encoder_Decoder}
\begin{figure}
	\centering
    \includegraphics[width=1\columnwidth, angle=0]{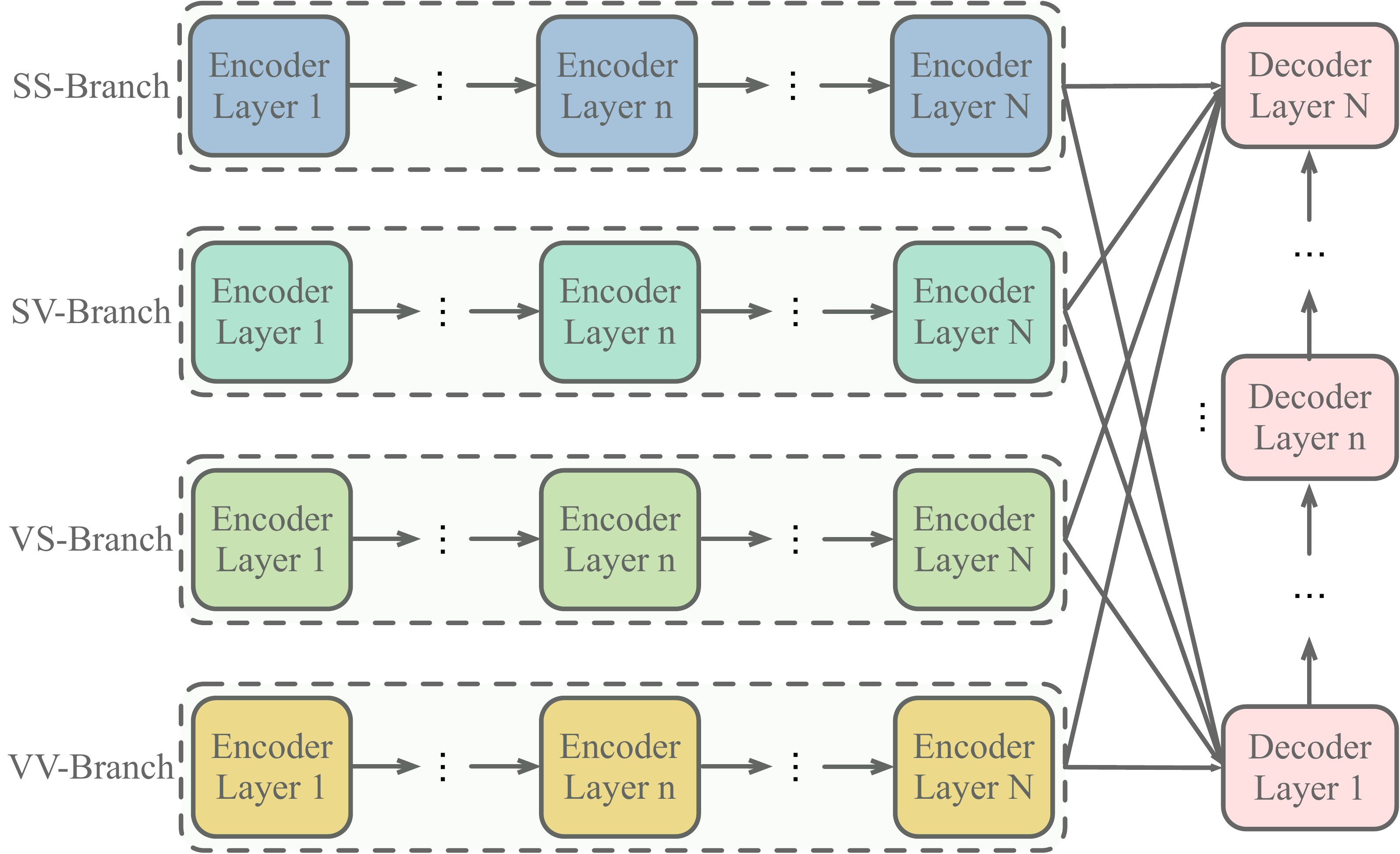}
	\vspace{-1ex}
	\caption{Our encoding part is composed of four branches($SS$ branch, $SV$ branch, $VS$ branch and $VV$ branch). 
	These encoders perform input's interaction with different visual-semantic ratios. Decoder is charge of generating textual tokens based on the weighted contribution of each branch.}
	\label{fig:Visual_Semantic_Branches}
\end{figure}

\noindent\textbf{Multi-Branches Encoder}. To bridge the modality gap, prior linguistic input are used to exploit semantic and visual information simultaneously.
However, the interaction between semantic and visual information is too complex to be encoded by just a weighted summation.
Thus, based on our fusion model, as shown is the Fig \ref{fig:Visual_Semantic_Branches},
four encoders are built parallelly, 
namely, $VV$(Pure Visual) branch, $VS$(Semantic fused to Visual) branch, $SV$(Visual fused to Semantic) branch and $SS$(Pure Semantic) branch.
In this way, model can not only mimic higher order visual-semantic interactions, 
but can also can generate a more fine-grained visual-semantic ratio.
The input-output relation in each branch is defined as
\begin{equation}
{\tilde{{X}}_{ss}} = \mathit{Branch_{ss}}({S},\\ {X_B^s}, \\ {G^{inter}_{s,v}},\\ {G^{intra}_{s,s}}),
\end{equation}
\begin{equation}
{\tilde{{X}}_{sv}} = \mathit{Branch_{sv}}({S},\\ {X_B^s},\\ {G^{inter}_{s,v}},\\ {G^{intra}_{s,s}}),
\end{equation}
\begin{equation}
{\tilde{{X}}_{vs}} = \mathit{Branch_{vs}}({V},\\ {X_B^v},\\ {G^{inter}_{v,s}},\\ {G^{intra}_{v,v}}),
\end{equation}
\begin{equation}
{\tilde{{X}}_{vv}} = \mathit{Branch_{vv}}({V},\\ {X_B^v},\\ {G^{inter}_{v,s}},\\ {G^{intra}_{v,v}}),
\end{equation}
where ${V}, {S}$ are visual feature and semantic dense-caption vectors. 
${X_B^v}, {X_B^s}$ are bounding box vectors for corresponding visual and semantic features.
${G^{inter}_{s,v(v,s)}}, {G^{intra}_{v,v(s,s)}}$ are inter-geometry and intra-geometry relation matrix for corresponding branches.
Therefore, 4 encoding branches will produce a multilevel output 
${\tilde{\chi}} = (\tilde{{X}}_{ss},\tilde{{X}}_{sv},\tilde{{X}}_{vs},\tilde{{X}}_{vv})$, 
obtained from the outputs of each branch.

\noindent\textbf{Multi-Input Decoder}. Conditioned on both previously generated words and encoder's output, 
decoder aims to generate the next tokens of the output caption. 
Just like the case in \cite{B_6}, here we also have multi-parallel encoder outputs,
Given an input sequence of vectors ${Y}$,
attention operator should connects ${Y}$ to all elements in ${\tilde{\chi}}$ through cross-attentions.
The decoder structure is basically the same to previous work, 
the only difference is that, the multiple inputs in \cite{B_6} are coming from different layers of the same encoder,
whereas, our inputs are coming from the last layers of different encoders.
Formally, our attention operator is similarly defined as
\begin{equation}
\mathit{Att}({\tilde{\chi}}, {Y}) = \sum_{i} \alpha_{i} \odot C(\tilde{{X}}_{i}, {Y}), i \in \{ss,sv,vs,vv\}
\end{equation}
\begin{equation}
\alpha_{i} =\theta({W_i}[{Y}, C(\tilde{{X}}_{i}, {Y})]+b_i)
\end{equation}
\begin{equation}
C(\tilde{{X}}_{i}, {Y}) = \mathit{Attention}({W_Q}{Y}, {W_K}\tilde{{X}}_{i}, {W_V}\tilde{{X}}_{i})
\end{equation}
where $C(.,.)$ stands for the encoder-decoder cross-attention, computed using queries from the decoder and keys and values from the encoder.
$[.,.]$ stands for the concatenation.
$\alpha_i$ is a matrix of weights having the same size as the cross-attention results. 
Weights in $\alpha_i$ modulate both the single contribution of each branch, 
and the relative importance between different branches. 
${W_i}$ is a $2d\times{d}$ weight matrix, and ${b_i}$ is a learnable bias vector.

In the testing stage, the caption is generated word-by-word in a sequential manner. 
When generating the $t$th word, the input features are represented as $Y_{\leq{t}} =[y_1,y_2,\cdots,y_{t-1}, {0}, \cdots, {0}] \in \mathbb{R}^{n\times{d_y}}$.
The input caption features along with the image features are fed forward the model to obtain the word with the largest probability among the whole word vocabulary.
The predicted word is then integrated into the inputs to recursively generate the new inputs $Y_{\leq{t+1}}$. 
To improve the diversity of generated captions, we also introduce the beam search strategy during the testing stage.

\section{Experiments}
\label{sec:Experiments}
In this section, we conduct experiments and evaluate the proposed GEVST models on MSCOCO 2015 image captioning dataset\cite{G_1} which is the most commonly used captioning dataset. 
Additionally, Visual Genome dataset\cite{G_3} is used to extract the bottom-up-attention visual features\cite{A_8} and also used to train the Dense-Caption model\cite{D_1}.

\begin{figure*}[ht]
	\includegraphics[width=2.05\columnwidth, angle=0]{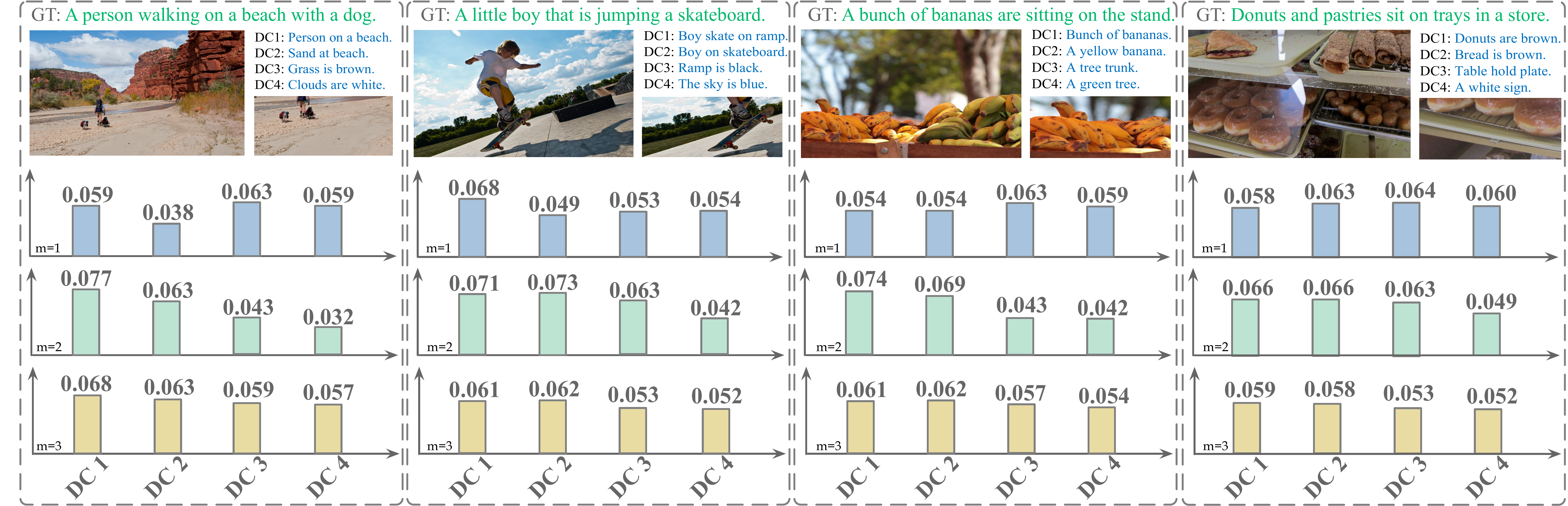}
	\vspace{-1ex}
	\caption{Visualization of the results with different cell numbers.
	For each original image, GT is the ground truth, DC$1$ to DC$4$ are 4 dense-captions for visualization.
	The sub-image is shown under the dense-captions.
	The number above the bar stand for the weight for the dense-captions(DC1 to DC4). $m$ stands for the number of fusion cells.}
	\label{fig:Experient_different_fusion_layers}
	\vspace{-0.0cm}
\end{figure*}

\begin{figure*}[ht]
	\includegraphics[width=2.05\columnwidth, angle=0]{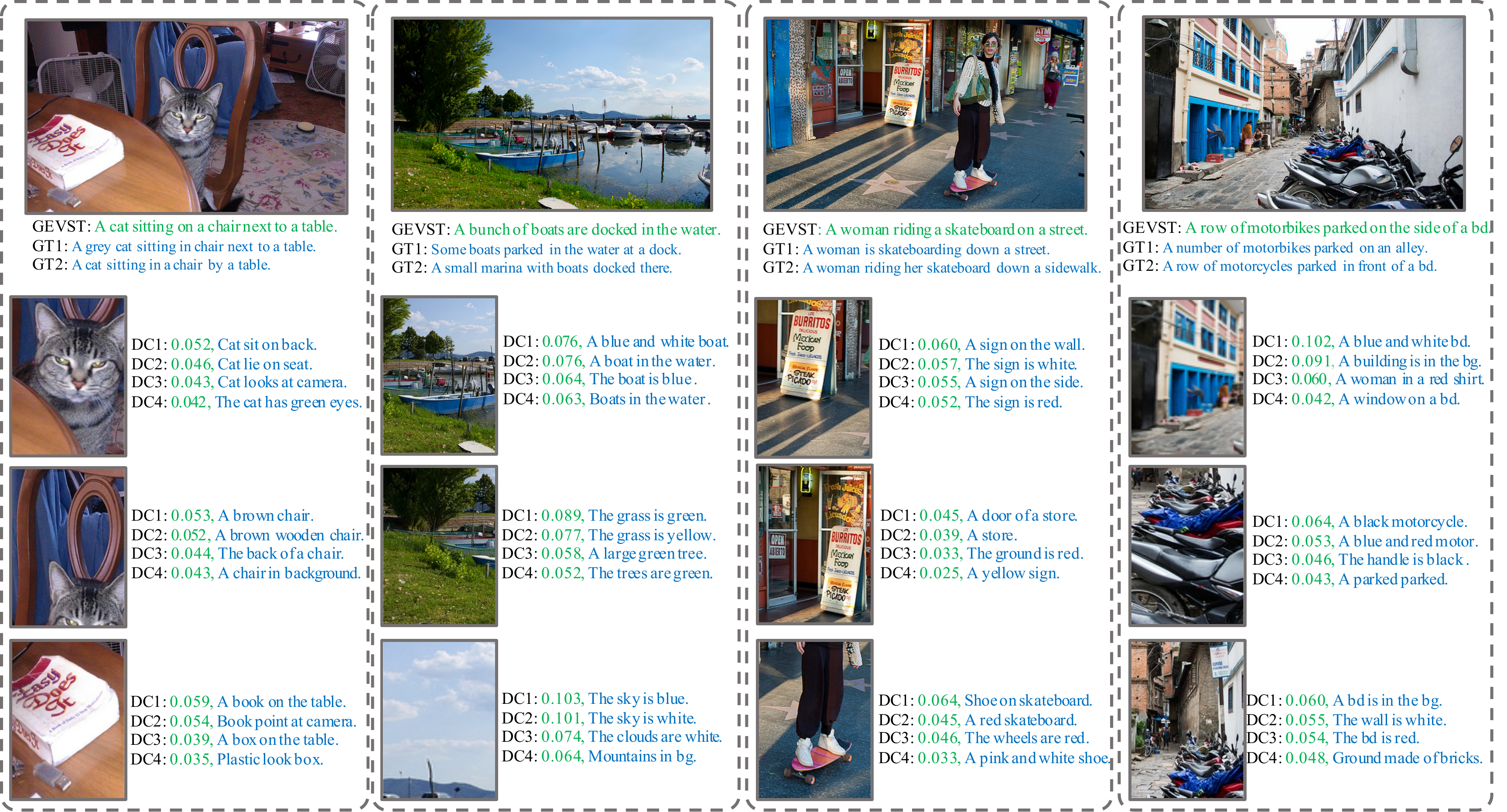}
	\vspace{-1ex}
	\caption{Visualization of the results with different sub-images.
	Under each original image, GT$1$ and GT$2$ are the ground truth captions. GEVST stands for the caption generated by our model.
	For a specific sub-image, DC$1$ to DC$4$ are 4 dense-captions with the highest attention weight.}
	\label{fig:generated_caption_and_fusion_attention}
	\vspace{-0.0cm}
\end{figure*}

\subsection{Datasets}
\label{sec:Data_Preparation}
\noindent\textbf{MSCOCO}\cite{G_1}. 
MSCOCO contains $164,062$ images with $995,684$ captions,
which are divided into three parts, the \emph{training} part ($82,783$ images),
the \emph{validation} part ($40,504$ images) and the \emph{test} part
($40,775$ images\footnote{Note that these images are unavailable and withheld in \textsf{MSCOCO} server for \emph{on-line} comparison.}.).
To evaluate the quality of the generated captions, we follow the
``\emph{karpathy split}'' evaluation strategy, $5,000$ images are chosen for \emph{offline} validation
and another $5,000$ images are selected for \emph{offline} testing.
To evaluate the quality of the generated captions, we adopt the widely used evaluation metrics,
\textbf{BLEU}\cite{F_4}, \textbf{METEOR}\cite{F_2}, \textbf{RougeL}\cite{F_3} and \textbf{CIDEr}\cite{F_4}.
Additionally, we use {\bf SPICE}\cite{F_1} to evaluate our model, which is more similar to human evaluation criteria.
For all indicators, the higher the better.

\noindent\textbf{Visual Genome}\cite{G_3}.
Visual Genome is a large-scale dataset to evaluate the interactions between objects in the images. 
For Dense-Caption training, similar to \cite{D_1}, $77,398$ images for training and $5000$ images each for validation and test.
We use the same evaluation metric of mean Average Precision (mAP) as \cite{D_1}, 
which measures localization and description accuracy jointly. 
For localization, IoU thresholds $.3, .4, .5, .6, .7$ are used. 
For language similarity, Meteor score thresholds $0, .05, .1, .15, .2, .25$ are used.

\subsection{Implementation Details}
\label{sec:Data_Preparation}
For the captions, we perform the pre-processing as follows. 
All the caption sentences are converted to lower case and tokenized into words with white space. 
The rare words that occur less than 5 times are discarded, 
resulting in a vocabulary of $9,487$ words. 
The out-of-vocabulary words are represented as $<$UNK$>$ token.
Corresponding word embeddings are trained with the model.

For the visual features, 
we use the pre-trained bottom-up-attention models to detect the objects and extract visual features for the detected objects. 
Each original region feature is a $2,048$-dimensional vector, 
which is transformed as the input visual feature with the dimension $d_o$=$512$.
For the semantic features,
Input dense-captions are encoded by a basic transformer-encoder with $3$ encoding layers.
For a dense-caption, the corresponding feature is the mean of encoder outputs.
Each original dense-caption feature is a $1,024$-dimensional vector, 
which is transformed as the input semantic feature with the dimension $d_o$=$512$.

Expansion rate $Er$ in fusion module is $5$. 
The number of Geometry Entangled Fusion blocks $m$ ranges in $\{1, 2, 3\}$.
The latent dimensionality $d_{model}$ in the Geometry-Entangle Self-Attention module is $512$,
the number of heads $h$ is $8$, and the latent dimensionality for each head $d_h$=$d/h$=$64$. 
The number of encoding layer $L$ in the encoder and decoder ranges in $\{2, 3, 4, 5\}$.

The whole image captioning architecture are mainly implemented with PyTorch, optimized with Adam\cite{E_4}. 
For the training stage, we follow the training schedule in\cite{D_8} to optimize the whole architecture with cross-entropy loss. 
The warmup steps are set as $4$ epochs and the batch size is $40$. 
For the training with self-critical training strategy,
we first select the initialization model which is trained with cross entropy loss for $18$ epochs. 
After that, the whole architecture is further optimized with CIDEr reward for another $30$ epochs.
At the inference stage, we adopt the beam search strategy and set the beam size as 5. 
Five evaluation metrics, BLEU@N, METEOR, ROUGE-L, CIDEr and SPICE, are simultaneously utilized to evaluate our model.

\subsection{Comparison with the State-of-the-Art}
\label{sec:Comparison_with_the_State-of-the-Art}
\begin{table}
    \centering
    \fontsize{8}{11}\selectfont
    \caption{Comparison results on MSCOCO on Karpathy's split. The heighest score is labbles as bold. Our model noted as GEVST.}
    \begin{tabular}{ccccccc}
    \toprule
    \toprule
    \textit{Model}& B-1 & B-4 & M & R & C & S \cr
                         
    \cmidrule(lr){1-7}
     SCST \tiny CVPR2017                 & -    & 34.2 & 26.7 & 57.7 & 114.0 & - \cr
     Up-Down \tiny CVPR2018              & 79.8 & 36.3 & 27.7 & 56.9 & 120.1 & 21.4 \cr
     GCN-LSTM \tiny ECCV2018             & 80.5 & 38.2 & 28.5 & 58.5 & 128.3 & 22.0 \cr
     HAN  \cite{H_1}\tiny AAAI2019       & 80.9 & 37.6 & 27.8 & 58.1 & 121.7 & 21.5 \cr
     SGAE \tiny CVPR2019                 & 80.8 & 38.4 & 28.4 & 58.6 & 127.8 & 22.1 \cr
     ORT \tiny NIPS2019                  & 80.5 & 38.6 & 28.7 & 58.4 & 127.8 & 22.1 \cr
     AoA\tiny ICCV2019                   & 80.2 & 38.9 & \bf{29.2} & 58.8 & 129.8 & 22.4 \cr
     HIP\tiny ICCV2019                   & -    & 39.1 & 28.9 & \bf{59.2} & 130.6 & 22.3 \cr
     SRT \cite{H_2}\tiny AAAI2020        & 80.3 & 38.5 & 28.7 & 58.4 & 129.1 & 22.4 \cr
     $M^{2}$ \tiny CVPR2020              & 80.8 & 39.1 & 29.2 & 58.6 & 131.2 & 22.6 \cr
    
    \cmidrule(lr){1-7}
    GEVST\tiny ours & \bf{81.1} & \bf{39.3} & 29.1 & 58.7 & \bf{131.6} & \bf{22.8} \cr
    \bottomrule
    \bottomrule
    \end{tabular}\vspace{0cm}
    \label{tab:Offline_MSCOCO}
\end{table}

\begin{table*}
    \centering
    \fontsize{8}{11}\selectfont 
    \caption{Leaderboard of the published state-of-the-art, $single-model$ methods on the online MS-COCO test server, where c5 and c40 denote
             using 5 and 40 references for testing, respectively. The highest score is denoted as bold. Our model is noted as GEVST.}
    \begin{tabular}{ccccccccccccccc}
    \toprule
    \toprule
    \multirow{2}{*}{\textit{Model}}&
    \multicolumn{2}{c}{B-1} & \multicolumn{2}{c}{B-2} & \multicolumn{2}{c}{B-3} & \multicolumn{2}{c}{B-4} & \multicolumn{2}{c}{M} & \multicolumn{2}{c}{R} & \multicolumn{2}{c}{C} \cr
    \cmidrule(lr){2-3} \cmidrule(lr){4-5} \cmidrule(lr){6-7} \cmidrule(lr){8-9} \cmidrule(lr){10-11} \cmidrule(lr){12-13} \cmidrule(lr){14-15}
    & c5 & c40 & c5 & c40 & c5 & c40 & c5 & c40 & c5 & c40 & c5 & c40 & c5 & c40 \cr
                         
    \cmidrule(lr){1-15}
     SCST \tiny CVPR2017                 & 78.1 & 93.7 & 61.9 & 86.0 & 47.0 & 75.9 & 35.2 & 64.5 & 27.0 & 35.5 & 56.3 & 70.7 & 114.7 & 116.7 \cr
     Stack-Cap \tiny AAAI2018            & 77.8 & 93.2 & 61.6 & 86.1 & 46.8 & 76.0 & 34.9 & 64.6 & 27.0 & 35.6 & 56.2 & 70.6 & 114.8 & 118.3 \cr
     Up-Down \tiny CVPR2018              & 80.2 & \bf{95.2} & 64.1 & 88.8 & 49.1 & 79.4 & 36.9 & 68.5 & 27.6 & 36.7 & 57.1 & 72.4 & 117.9 & 120.5 \cr
     CVAP \tiny MM2018                   & 80.1 & 94.9 & 64.7 & 88.8 & 50.0 & 79.7 & 37.9 & 69.0 & 28.1 & 37.0 & \bf{58.2} & 73.1 & 121.6 & 123.8 \cr
     HAN  \tiny AAAI2019       & 80.4 & 94.5 & 63.8 & 87.7 & 48.8 & 78.0 & 36.5 & 66.8 & 27.4 & 36.1 & 57.3 & 71.9 & 115.2 & 118.2 \cr
     SGAE \tiny CVPR2019                 & 80.6 & 95.0 & 65.0 & 88.9 & 50.1 & 79.6 & 37.8 & 68.7 & 28.1 & 37.0 & \bf{58.2} & 73.1 & 122.7 & 125.5 \cr
     VSUA \cite{H_3} \tiny MM2019                   & 79.9 & 94.7 & 64.3 & 88.6 & 49.5 & 79.3 & 37.4 & 68.3 & 28.2 & 37.1 & 57.9 & 72.8 & 123.1 & 125.5 \cr
    
    \cmidrule(lr){1-15}
    GEVST\tiny ours & \bf{80.8} & 95.1 & \bf{65.1} & \bf{89.2} & \bf{50.4} & \bf{80.5}  & \bf{38.2} & \bf{70.1} & \bf{28.7} & \bf{37.9} & \bf{58.2} & \bf{73.3} & \bf{125.1} & \bf{127.8} \cr
    
    \bottomrule
    \bottomrule
    \end{tabular}\vspace{0cm}
    \label{tab:Online_MSCOCO}
\end{table*}

\noindent\textbf{Offline Evaluation}.
As shown in Table \ref{tab:Offline_MSCOCO} We report the performance on the offline test split of our model as well as the compared models.
All scores are tested on single model for comparison.
The models include: 
SCST~\cite{E_1}, which employs Reinforcement Learning after XE training.
Up-Down~\cite{A_8}, which employs a two LSTM layers model with bottom-up features extracted from Faster-RCNN.
GCN-LSTM~\cite{C_5}, which builds graphs over the detected objects in an image based on their spatial and semantic connections.
HAN~\cite{B_7}, constructs a feature pyramid leveraging patch features and also a parallel Multivariate Residual Network to integrate features of different levels.
SGAE~\cite{C_6}, which introduces auto-encoding scene graphs into its model.
ORT~\cite{B_8}, which uses bounding boxes to fortify position relation representation.
AoA~\cite{B_4}, which modifies traditional attention model by adding an additional weight gate.
HIP~\cite{H_1}, which uses tree-structured Long Short-Term Memory network to interpret the hierarchal structure and enhance 3 levels' features.
SRT~\cite{H_2}, which applies an image-text matching model to retrieve recalled words for each image and also two methods to utilize these recalled words.
$M^{2}$~\cite{B_6}, which proposes a multi-layer encoder for image regions and a multi-layer decoder. Also, extra slots are used to learn priori knowledge.
For fair comparison, all the models are first trained under XE loss and then optimized for CIDEr-D score.

\noindent\textbf{Online Evaluation}.
We also evaluate our model on the online COCO test server 
Finally, we also report the performance of our method on the online COCO test server as shown in Table \ref{tab:Online_MSCOCO}.
In this case, we still use the single models previously described, trained on the “Karpathy” training split. 
The evaluation is done on the COCO test split, for which ground-truth annotations are not publicly available. 
In comparison with the state-of-the-art approaches of the leaderboard.
As it can be seen, our model achieves the highest scores for most metrics. 
For fair comparison, we only list models which have online socres of single model.

\subsection{Qualitative Analysis}
\label{sec:Qualitative_Analysis}
To better understand the effectiveness of proposed approach and the explanations mentioned above,
We visualize the dense-caption attention weights with different fusion cell numbers as shown in Fig \ref{fig:Experient_different_fusion_layers}.
dense-caption attention weights with different sub-images are also shown in the Fig \ref{fig:generated_caption_and_fusion_attention}.

\noindent\textbf{Number of Fusion Cells}.
Instead of a single cell,
a stack structure can fortify the effectiveness of fusion, which can improve the model's performance.
We can see from the Fig \ref{fig:Experient_different_fusion_layers},
compare to $m=1$, model with $m=2$ gives more reasonable attention weights.
Whereas, increasing fusion cell number doesn’t always mean better performance.
When we set $m$ to $3$, the attention weights become similar to each others,
fusion model fails to distinguish the most important dense-caption for the given sub-image.

\noindent\textbf{Effective of Visual-Semantic Fusion}.
Fig \ref{fig:generated_caption_and_fusion_attention} shows some examples of fusion attention. 
From these examples, we find that our fusion model has three superior abilities:
Firstly, compare to original image, our fusion model can provide more diversity info. 
For example, as for the cat in the second row, model can provide richer descriptions. 
Whereas, previous works fail to give such fine grained level description.
Secondly, model is able to discriminates the importance of objects. 
Since, for objects with less importance, fewer related dense-captions are provided by our model. 
For example, comparing the cats and the buildings sub-image in the fourth row, 
We can get more dense-captions about the cat. In contrast, fewer dense-captions are generated for the background buildings.
Lastly, model can describe objects relation more accurately.
Rather than reasoning from the visual information monopoly,
model can get Richer and more accurate relationships from dense-captions straightforwardly.

\subsection{Ablation Studies}
\label{sec:Ablation_Studies}
We run a number of ablation experiments on MSCOCO image captioning dataset to explore the effectiveness of GEVST model with different hyperparameters. 
The results shown in the following tables which are discussed in detail below.

\noindent\textbf{Geometry-Content Fusion}.
\begin{table}
    \centering
    \fontsize{8}{11}\selectfont 
    \caption{Ablation studies for fusion scheme: 
             Scores of different fusion schemes $m-B$, where $m \in \{1,2,3\}$ means the number of fusion cells, 
             $B \in \{C,G,CG\}$ means which kind of base is used by fusion cells.
             ($C$:Only Content Based, $G$:Only Geometry Based, $CG$:Both Content and Geometry Based).}
    \begin{tabular}{ccccccccc}
    \toprule
    \toprule
    \multirow{2}{*}{\textit{Scheme}}&
    \multicolumn{4}{c}{X-Entropy(30 Epoch)} & \multicolumn{4}{c}{CIDEr Optimization} \cr
             \cmidrule(lr){2-5}  
             \cmidrule(lr){6-9}
                                & B-4 & M & R & C & B-4 & M & R & C \cr

    \cmidrule(lr){1-9}
     1-CG & 36.6 & 28.3 & 57.2 & 118.0 & 37.3 & 28.8 & 58.4 & 128.6 \cr
     2-CG & \bf{37.0} & \bf{28.4} & \bf{57.3} & \bf{118.3} & \bf{39.3} & \bf{29.1} & \bf{58.7} & \bf{131.6} \cr
     3-CG & 36.8 & 28.2 & \bf{57.3} & 117.7 & 38.6 & 29.0 & 58.3 & 129.7 \cr

    \cmidrule(lr){1-9}
     2-C     & 36.9 & 28.3 & 57.2 & 118.1 & 38.7 & 29.0 & 57.3 & 127.1 \cr
     2-G     & 36.3 & 28.1 & 56.9 & 116.2 & 37.6 & 28.7 & 56.6 & 124.3 \cr
     2-CG    & \bf{37.0} & \bf{28.4} & \bf{57.3} & \bf{118.3} & \bf{39.3} & \bf{29.1} & \bf{58.7} & \bf{131.6} \cr

    \bottomrule
    \bottomrule
    \end{tabular}\vspace{0cm}
    \label{tab:Ablation_Study_Fusion}
\end{table}

Table \ref{tab:Ablation_Study_Fusion} summarizes the ablation experiments on different fusion schemes for Stack Geometry-Entangled Fusion.
The first parameter we changed is the cell number $m$ in the fusion model,
we set the cell number from $1$ to $3$,
Regarding the performance, 
the model’s performance improves as we increase $m$ to $2$.
That means single fusion cell can hardly produce the precise alignment attention between two modalities,
an extra fusion cell can help model achieves a preciser alignment.
However, for $3$-rd fusion cell, the queries are fused twice from previous fusion cells,
thus, attention scores are too general to distinguish the difference between input features.
The second parameter we investigated is the Fusion-Base,
it decides which kind of information guides the alignment between two modalities features.
We can see, when we use content information and geometry information simultaneously, 
model gets the best performance.
This is because, for some objects, their relationships are decided by the geometry information($eg$: Planes above the bridge.). 
However, for other objects, their relationships are decided by their content information($eg$: Girl eats cake.).
Relying on a single kind of relation can't deal with all cases mentioned above.

\noindent\textbf{Number of Attention Layers}.
\begin{table}
    \centering
    \fontsize{8}{11}\selectfont
    \caption{Ablation studies for number of Self-Attention layers: 
             Scores of the models with different attention layer numbers(Encoder and Decoder have the same layer numbers) $l \in \{2,3,4,5\}$.}
    \begin{tabular}{ccccccccc}
    \toprule
    \toprule
    \multirow{2}{*}{\textit{l}}&
    \multicolumn{4}{c}{X-Entropy(30 Epoch)} & \multicolumn{4}{c}{CIDEr Optimization} \cr
             \cmidrule(lr){2-5}  
             \cmidrule(lr){6-9}
                                & B-4 & M & R & C & B-4 & M & R & C \cr

    \cmidrule(lr){1-9}
     2 & \bf{37.0} & 28.2 & 57.0 & 118.2 & 38.5 & 28.4 & 57.5 & 128.1 \cr
     3 & \bf{37.0} & \bf{28.4} & \bf{57.3} & \bf{118.3} & \bf{39.3} & \bf{29.1} & \bf{58.7} & \bf{131.6} \cr
     4 & 36.8 & 28.2 & \bf{57.3} & 117.2 & 38.6 & 28.7 & 58.0 & 129.6 \cr
     5 & 36.1 & 27.9 & 56.6 & 114.1 & 38.3 & 28.2 & 57.0 & 128.0 \cr

    \bottomrule
    \bottomrule
    \end{tabular}\vspace{0cm}
    \label{tab:Ablation_Study_Layer_Number}
\end{table}

Table \ref{tab:Ablation_Study_Layer_Number} shows the performance of 
the GEVST model with different number of attention layers. 
Regarding the performance,
as increasing $l$, the model’s performance gradually improves 
and is saturated at a certain number. 
This can be explained as a deeper model can capture more complex relationships among objects, 
providing a more accurate understanding of the image contents. 
In addition, a deeper model has a larger representation capacity and has a larger risk to overfit the training set, 
thus, the performance gets worse when setting the $l$ too large.

\noindent\textbf{Geometry-Entangled Self-Attention}.
\begin{table}
    \centering
    \fontsize{8}{11}\selectfont
    \caption{Ablation studies for number of Geometry-Entangled Self-Attention: 
             Scores of the models with different Self-Attention module, the basic module just use content information, 
             $+X$ means module $X$ are added to the previous row. $X$ $\in$ \{\mbox{Intra Geometry(Intra)}, \mbox{Inter Geometry(Inter)}\}.}
    \begin{tabular}{ccccccccc}
    \toprule
    \toprule
    \multirow{2}{*}{\textit{X}}&
    \multicolumn{4}{c}{X-Entropy(30 Epoch)} & \multicolumn{4}{c}{CIDEr Optimization} \cr
             \cmidrule(lr){2-5}  
             \cmidrule(lr){6-9}
                                & B-4 & M & R & C & B-4 & M & R & C \cr

    \cmidrule(lr){1-9}
     Con   & 36.1      & 28.1      & 56.8      & 113.2 & 37.9 & 28.4 & 56.4 & 128.6 \cr
     +Intra & \bf{37.1} & \bf{28.4} & 57.2      & 117.8 & 38.5 & 28.7 & 57.7 & 129.7 \cr
     +Inter & 37.0      & \bf{28.4} & \bf{57.3} & \bf{118.3} & \bf{39.3} & \bf{29.1} & \bf{58.7} & \bf{131.6} \cr

    \bottomrule
    \bottomrule
    \end{tabular}\vspace{0cm}
    \label{tab:Geometry-Entangled Self-Attention}
\end{table}

Next, we compare the GEVST model with different Self-Attention modules in Table \ref{tab:Geometry-Entangled Self-Attention}. 
From the results,
we can see following that: 
1) By adding more geometry information to the Self-Attention module, 
the performance gets better steadily, 
thus highlighting the effect of using geometry information;
and 2) The optimal model is trained with both intra- and inter-geometry information, 
thus inter-geometry information is proved to be effective in attention measurement refinement.
For the cross-entropy loss, the performance of $+Intra$ is very similar to $+Inter$,
but when trained with reinforcement learning, $+Inter$ perform the best.
This can be explained as the reinforcement learning-based self-critical loss provides a more diverse exploration 
of the hypothesis space to avoid overfitting, 
and thus it can better utilize the potential of more complex models.

\noindent\textbf{Multi-Branch Encoder}.
\begin{table}
    \centering
    \fontsize{8}{11}\selectfont
    \caption{Ablation studies for number of Encoder-Branch: 
             Scores of the models with different Encoders. $VV$ means visual encoder, $VS$ means visual-semantic encoder(semantic features are fused into visual features), $SV$ means semantic-visual encoder(visual features are fused into semantic features), $SS$ means semantic encoder. $+X$ means encoder branch $X$ is added to the model of last row.}
    \begin{tabular}{ccccccccc}
    \toprule
    \toprule
    \multirow{2}{*}{\textit{Encoder}}&
    \multicolumn{4}{c}{X-Entropy(30 Epoch)} & \multicolumn{4}{c}{CIDEr Optimization} \cr
             \cmidrule(lr){2-5}  
             \cmidrule(lr){6-9}
                                & B-4 & M & R & C & B-4 & M & R & C \cr
                         
    \cmidrule(lr){1-9}
     SS    & 31.3 & 22.3 & 51.7 & 101.5 & 35.2 & 25.9 & 54.2 & 116.9 \cr
     SV    & 33.2 & 24.1 & 52.9 & 104.9 & 35.9 & 26.2 & 54.8 & 118.9 \cr
     VS    & 36.9 & 28.1 & 57.0 & 117.1 & 38.4 & 28.6 & 57.4 & 128.3 \cr
     VV    & 36.1 & 27.7 & 56.6 & 114.1 & 37.2 & 27.6 & 56.9 & 125.3 \cr
     +VS   & \bf{37.2} & 28.3 & 57.2 & 117.9 & 38.9 & 28.6 & 57.7 & 128.4 \cr
     +SV   & \bf{37.2} & \bf{28.4} & 57.2 & \bf{118.5} & 39.2 & \bf{29.1} & 58.0 & 129.4 \cr
     +SS   & 37.0 & 28.4 & \bf{57.3} & 118.3 & \bf{39.3} & \bf{29.1} & \bf{58.7} & \bf{131.6} \cr

    \bottomrule
    \bottomrule
    \end{tabular}\vspace{0cm}
    \label{tab:Geometry-Entangled-Branch}
\end{table}

In Table \ref{tab:Geometry-Entangled-Branch}, 
we show the performance of the GEVST model with different encoders. 
Comparing models with single branch(row-$1$ to row-$4$),
We can see that: 
1) model with fused input features perform better than those only use single modality features.
which justified that providing semantic information can narrow the gap in these two modalities.
2) model with visual-based information perform better than those semantic-based models.
Since dense-captions are homologous to the captions we want to generate, 
the errors in these dense-captions will also be transfer to the decoder directly,
which deteriorate the performance.
Besides,
by comparing models form row-$4$ to row-$7$, we can see that:
as we add more encoders, the performance gets better steadily.
This justifies that effectiveness of every branch we add.
For training using the cross-entropy loss, 
the optimal model is the one with $3$ encoders ($VV$,$VS$,$SV$).
This phenomenon is due to the fact that larger models need more time to converge,
since, the comparison is done by 30 epochs, 
we infer that model with $4$ encoders ($VV$, $VS$, $SV$, $SS$) can achieve better scores with a longer training time.
Its capability is demonstrated in the case training with CIDEr optimization.

\section{Conclusion}
\label{sec:Conclusion}
In this paper, we present a novel Geometry-Entangled Visual Semantic Transformer (GEVST) framework for image captioning. 
Our GEVST consists of four parallel transformer encoders VV(Pure Visual), VS(Semantic fused to Visual), 
SV(Visual fused to Semantic), SS(Pure Semantic) for final caption generation.
To take fully advantage of the capacity of dense-caption semantic features,
Both content and geometry relationships of visual and semantic features are considered in fusion model and self-attention encoder.
Specially, the geometry fusion attention will not only refine the fusion attention measurement, 
but also be responsible for inter-geometry feature generation.
We quantitatively and qualitatively evaluate the proposed GEVST models on the benchmark MSCOCO image captioning dataset.
Extensive ablation studies are explored to find the reasons behind the GEVST's effectiveness. 
Experimental results show that our method significantly outperforms existing approaches, 
and for single model, GEVST achieves the best performance on the real-time leaderboard of the MSCOCO image captioning challenge.


%

\section*{Acknowledgment}
This work was supported in part by the National Natural Science Foundation of China under Grant 61602197, and in part by the Cognitive Computing and Intelligent Information Processing (CCIIP) Laboratory, Huazhong University of Science and Technology, Wuhan, China.

\ifCLASSOPTIONcaptionsoff
  \newpage
\fi



%

\bibliographystyle{IEEEtran}
\bibliography{reference}

%

\begin{IEEEbiography}[{\includegraphics[width=1in,height=1.25in,clip,keepaspectratio]{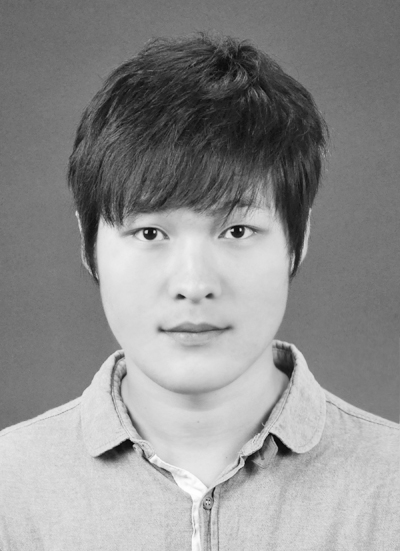}}]
    {LING CHENG} received the master degree from Fudan University, Shanghai, China, in 2019. He is currently a Ph.D. student at School of Computing and Information Systems, Singapore Management University, Singapore. His research interests include computer vision and data mining.
\end{IEEEbiography}

\begin{IEEEbiography}[{\includegraphics[width=1in,height=1.25in,clip,keepaspectratio]{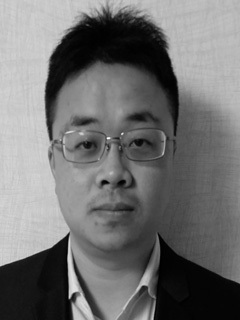}}]
	{WEI WEI} received the PhD degree from Huazhong University of Science and Technology, Wuhan, China, in 2012. He is currently an associate professor with the School Computer Science and Technology, Huazhong University of Science and Technology. He was a research fellow with Nanyang Technological University, Singapore, and Singapore Management University, Singapore. His current research interests include computer vision, natural language processing, information retrieval, data mining, and social computing.
\end{IEEEbiography}

\begin{IEEEbiography}[{\includegraphics[width=1in,height=1.25in,clip,keepaspectratio]{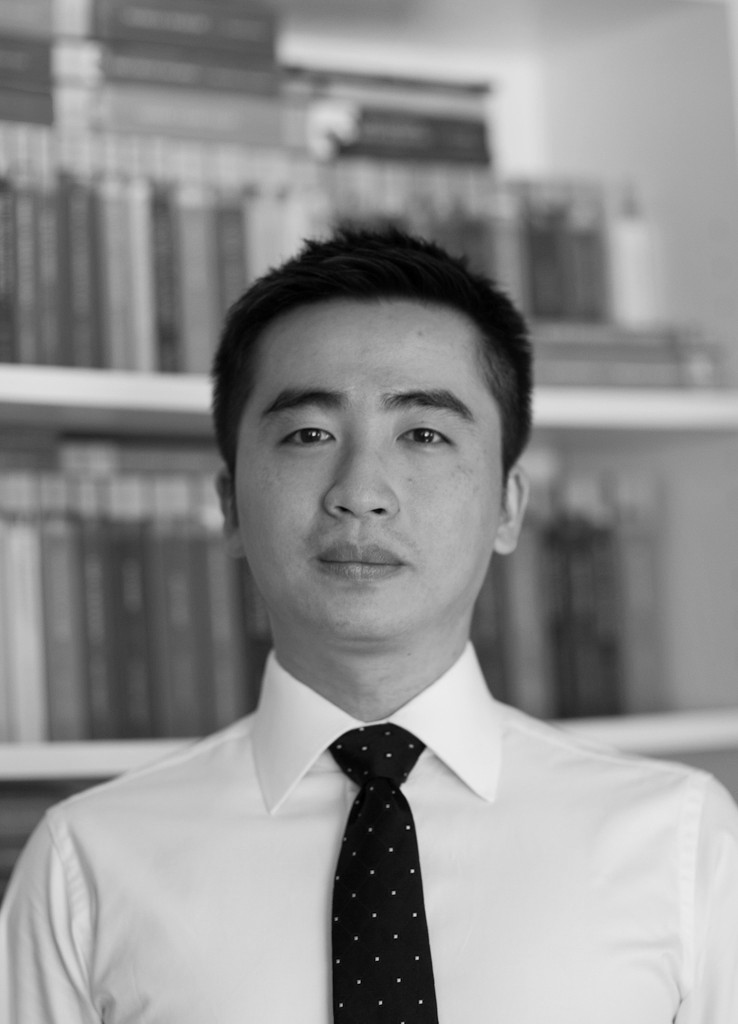}}]
	{FEIDA ZHU} received the B.S. degree in computer science from Fudan University, Shanghai, China, and the Ph.D. degree in computer science from
    the University of Illinois at Urbana–Champaign, Champaign, IL, USA. He is an Associate Professor with the School of Computing and Information Systems, Singapore Management University, Singapore. His current research interests include large-scale graph pattern mining and data mining, with applications on Web, management information systems and business intelligence.
\end{IEEEbiography}

\begin{IEEEbiography}[{\includegraphics[width=1in,height=1.25in,clip,keepaspectratio]{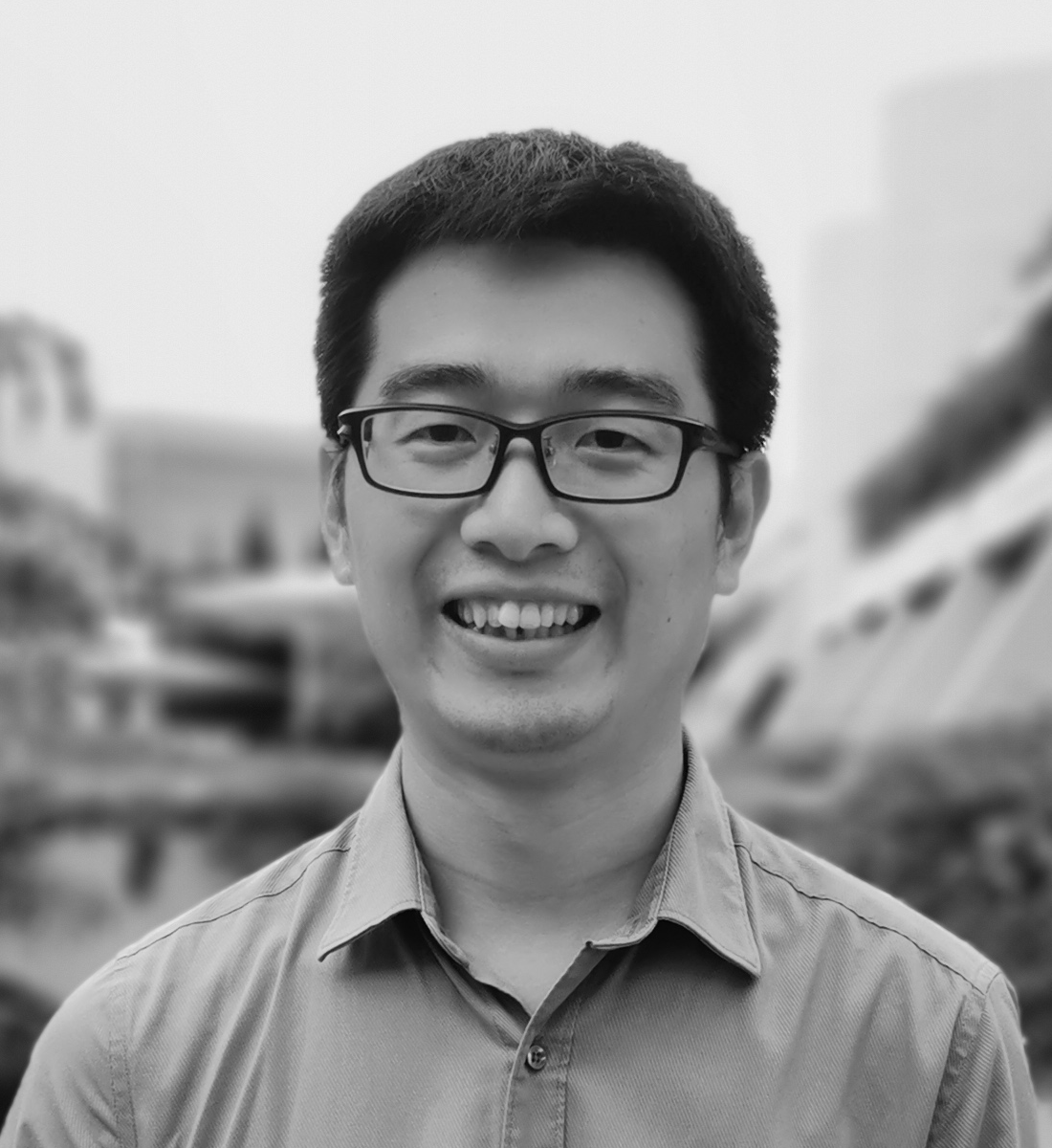}}]
	{YONG LIU} is a Research Scientist at Alibaba-NTU Singapore Joint Research Institute and Joint NTU-UBC Research Centre of Excellence in Active Living for the Elderly (LILY), Nanyang Technological University, Singapore. Prior to that, he was Data Scientist at NTUC Enterprise, Singapore from November 2017 to July 2018, and a Research Scientist at Data Analytics Department, Institute for Infocomm Research (I2R), A*STAR, Singapore from November 2015 to October 2017. He received Ph.D. from the School of Computer Science and Engineering at Nanyang Technological University in 2016 and B.S. from the Department of Electronic Science and Technology at University of Science and Technology of China in 2008. His research areas include various topics in machine learning and data mining. His research papers appear in leading international conferences and journals. He has been invited as a PC member of major conferences such as KDD, SIGIR, IJCAI, AAAI, CIKM, ICDM, and reviewer for IEEE/ACM transactions. He is a member of ACM and AAAI.
\end{IEEEbiography}

\begin{IEEEbiography}[{\includegraphics[width=1in,height=1.25in,clip,keepaspectratio]{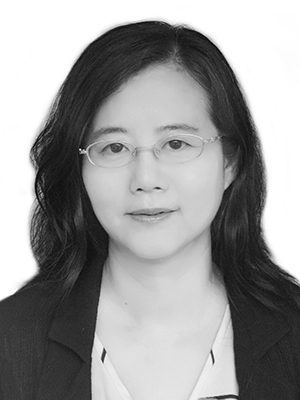}}]
	{CHUNYAN MIAO} is the chair of School of Computer Science and Engineering in Nanyang Technological University (NTU), Singapore. Dr. Miao is a Full Professor in the Division of Information Systems and Director of the Joint NTU-UBC Research Centre of Excellence in Active Living for the Elderly (LILY), School of Computer Engineering, Nanyang Technological University (NTU), Singapore. She received her PhD degree from NTU and was a Postdoctoral Fellow/Instructor in the School of Computing, Simon Fraser University (SFU), Canada. She visited Harvard and MIT, USA, as a Tan Chin Tuan Fellow, collaborating on a large NSF funded research program in social networks and virtual worlds. She has been an Adjunct Associate Professor/Associate Professor/Founding Faculty member with the Center for Digital Media which is jointly managed by The University of British Columbia (UBC) and SFU. Her current research is focused on human-centered computational/ artificial intelligence and interactive media for the young and the elderly. Since 2003, she has successfully led several national research projects with a total funding of about 10 Million dollars from both government agencies and industry, including NRF, MOE, ASTAR, Microsoft Research and HP USA. She is the Editor-in-Chief of the International Journal of Information Technology published by the Singapore Computer Society.
\end{IEEEbiography}




\end{document}